\documentclass[conference]{IEEEtran}
\IEEEoverridecommandlockouts
\usepackage{cite}
\usepackage{amsmath,amssymb,amsfonts}
\usepackage{algorithmic}
\usepackage{graphicx}
\usepackage{textcomp}
\usepackage{xcolor}
\usepackage{booktabs}
\usepackage{pifont}
\usepackage{url}
\usepackage{comment}
\usepackage{placeins}
\usepackage{float}

\def\BibTeX{{\rm B\kern-.05em{\sc i\kern-.025em b}\kern-.08em
T\kern-.1667em\lower.7ex\hbox{E}\kern-.125emX}}
\begin{document}
\bstctlcite{IEEEtranBSTCTL:disable_dash}

\title{Learning Emotion from Motion: Kinetic Multi-Stream Skeleton Modeling with Metadata-Conditioned Weak Label Distributions
}
\author{\IEEEauthorblockN{Anonymous Author}
}

\author{\IEEEauthorblockN{Sosuke Suzuki}
  \IEEEauthorblockA{\textit{University of Toyama} \\
    \textit{Department of Engineering}\\
    Toyama-city, Japan \\
  s2370241@ems.u-toyama.ac.jp}
  \and
  \IEEEauthorblockN{Yijin Wei}
  \IEEEauthorblockA{\textit{University of Toyama} \\
    \textit{Grad. Sch. of Science \& Engineering}\\
    Toyama-city, Japan \\
  d26c2003@ems.u-toyama.ac.jp}
  \and
  \IEEEauthorblockN{Koichiro Kamide}
  \IEEEauthorblockA{\textit{University of Toyama} \\
    \textit{Grad. Sch. of Science \& Engineering}\\
    Toyama-city, Japan \\
  m25c1014@ems.u-toyama.ac.jp}
  \and
  \IEEEauthorblockN{Ran Dong}
  \IEEEauthorblockA{\textit{Chukyo University} \\
    \textit{Faculty of Engineering}\\
    Nagoya-city, Japan \\
  randong@sist.chukyo-u.ac.jp}
  \and
  \IEEEauthorblockN{Haoran Xie}
  \IEEEauthorblockA{\textit{Japan Advanced Institute of Science and Technology} \\
    \textit{Faculty of Engineering}\\
    Nomi-city, Japan \\
  xie@jaist.ac.jp}
  \and
  \IEEEauthorblockN{Chao Zhang\thanks{Corresponding author: Chao Zhang (zhang@eng.u-toyama.ac.jp).}}
  \IEEEauthorblockA{\textit{University of Toyama} \\
    \textit{Faculty of Engineering}\\
    Toyama-city, Japan \\
  zhang@eng.u-toyama.ac.jp}
}
\maketitle

\begin{abstract}
  Skeleton-based emotion recognition from body motion remains challenging because emotional expressions are often characterized by subtle dynamic and relational motion cues, and hard labels may not fully capture ambiguity among related emotion categories. For the DIEM-A task in the MMAC ACII 2026 Challenge, we propose a multi-branch skeleton-based emotion recognition framework that combines a 6D rotation-based branch, a part-aware kinetic multi-stream branch, and a metadata-conditioned weak label distribution learning (LDL) branch. The branches are trained independently and fused by a probability-level ensemble at inference time. In 10-fold leave-performer-out cross-validation, the proposed framework improves Accuracy from 0.271 to 0.366 and Macro-F1 from 0.252 to 0.353 over the rotation-based baseline. Explainability ablations show that velocity and bone streams, as well as arm and leg regions, provide important cues for recognizing emotional body motion.
\end{abstract}


\begin{figure*}[t]
  \centering
  \includegraphics[width=0.90\linewidth]{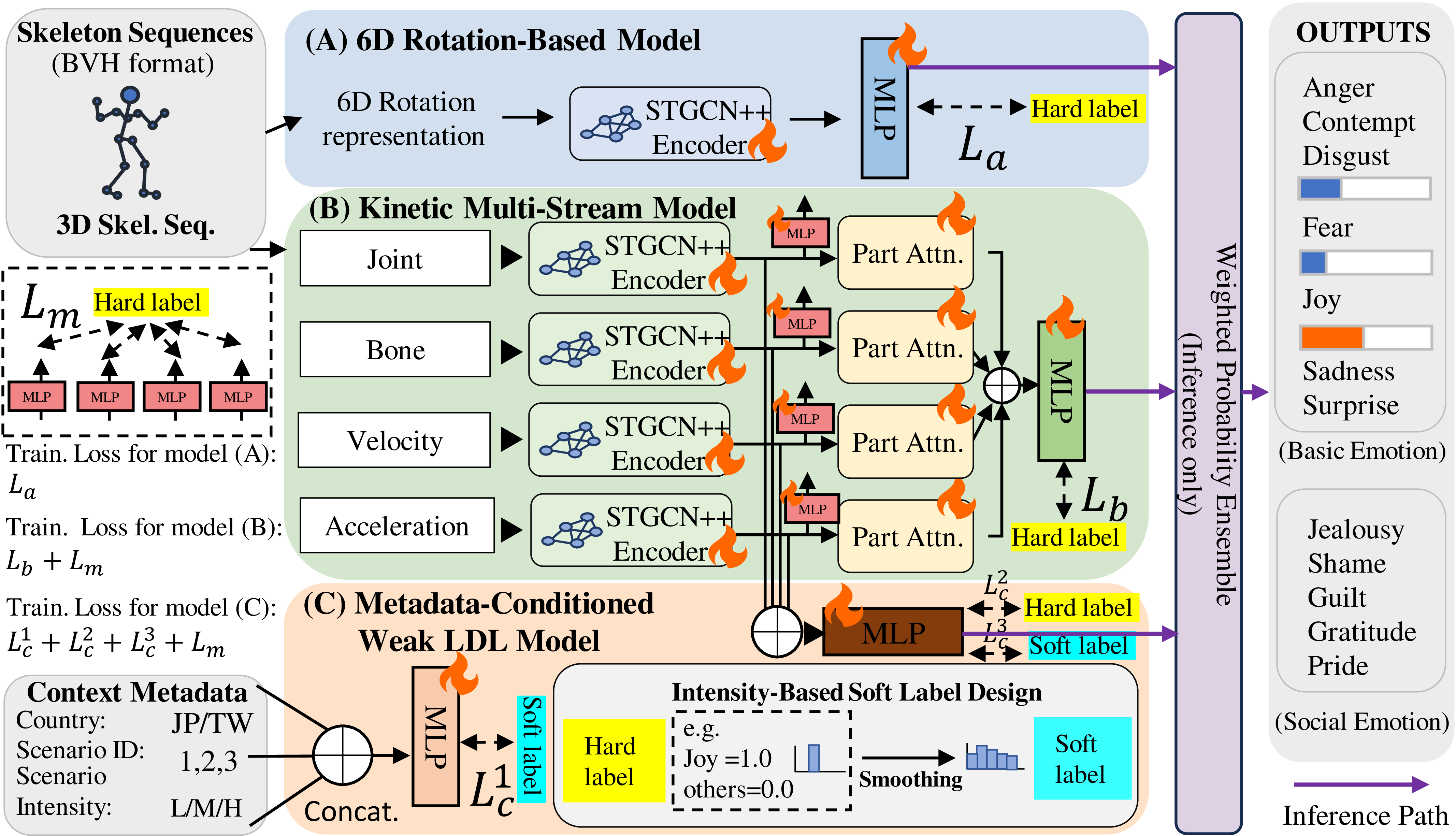}
  \caption{
    Overview of the proposed multi-branch skeleton-based emotion recognition framework.
    The framework consists of three independently trained branches: (A) 6D rotation-based branch, (B) part-aware kinetic multi-stream branch, and (C) metadata-conditioned weak LDL branch. During training, each branch is optimized with its own loss function, where (A) uses hard-label supervision, (B) uses hard-label supervision with stream-level auxiliary losses, and (C) combines the metadata auxiliary LDL loss \(L_c^1\), hard-label classification loss \(L_c^2\), motion-level LDL loss \(L_c^3\), and multi-stream auxiliary loss \(L_m\). During inference, the softmax probabilities from the three branches are combined by a probability-level ensemble to produce the final emotion prediction.
  }
  \label{fig:overview}
\end{figure*}

\section{Introduction}
\label{sec:introduction}

Human emotions are expressed not only through facial expressions and speech, but also through full-body movements, which provide important cues for affective computing and body gesture emotion recognition~\cite{noroozi2018emotionalbodygesture,piana2014bodygestures}.
Skeleton-based emotion recognition offers a compact and privacy-preserving way to model such body motion and has become an emerging direction in affective computing~\cite{lu2025skeletonemotion}.

In this work, we address the DIEM-A skeleton-based emotion recognition task~\cite{cheng2025diema} in the MMAC ACII 2026 Challenge~\cite{mmacacii2026challenge}. Each input is a BVH-format 3D skeleton sequence, and the goal is to classify it into one of 12 basic or social emotion categories.
Unlike action recognition, where motion patterns often correspond to explicit physical actions, emotional body expressions are often characterized by subtle motion cues and ambiguous boundaries between related emotion categories. The same emotion may appear with different movements, while different emotions may share similar motion patterns.

Skeleton-based human motion modeling has been widely applied to action recognition, anomaly detection, and motion prediction~\cite{wei2026human,kamide2025few,gu2024orientation}, demonstrating its effectiveness for learning structured human movement representations. Existing skeleton modeling frameworks, including ST-GCN-based models and rotation-based representations~\cite{yan2018stgcn,duan2022pyskl,zhou2019rotation6d}, are effective for modeling structured body motion.
However, when used alone, rotation-based representations do not explicitly encode translational and kinetic cues such as joint displacement, limb velocity, acceleration, and body expansion or contraction. These cues can be important for emotional body expressions, where affective states are often reflected in dynamic and relational kinematic patterns~\cite{piana2014bodygestures,shi2019twostreamagcn,shi2020msaagcn}.
Therefore, we argue that skeleton-based emotion recognition can benefit from complementary motion representations.

To this end, we propose a multi-branch framework consisting of three independently trained branches, as illustrated in Fig.~\ref{fig:overview}. Branch A is a 6D rotation-based STGCN++ branch that follows the official baseline setting. Branch B is a part-aware kinetic multi-stream branch using joint, bone, velocity, and acceleration streams from 3D joint positions. Branch C provides weak regularization for label ambiguity by training with intensity-aware weak soft-label distributions and metadata-based auxiliary supervision, rather than relying solely on hard one-hot labels.
At inference time, the three branches are fused only through a probability-level ensemble. This design avoids joint optimization while allowing rotation-based, kinetic, and weak ambiguity-regularized branches to provide complementary predictions. We also analyze kinematic stream and body-part importance to examine which motion cues and body regions contribute to the final predictions.
Our main contributions are summarized as follows:
\begin{itemize}
  \item We build a multi-branch ensemble framework for the DIEM-A skeleton-based emotion recognition task, combining rotation-based, kinetic multi-stream, and metadata-conditioned branches at the probability level.

  \item We introduce a part-aware kinetic multi-stream branch that integrates joint, bone, velocity, and acceleration streams with body-part encoding to capture complementary full-body emotion cues.

  \item We propose a metadata-conditioned weak label distribution learning (LDL) branch to address ambiguity among related emotion categories and analyze kinematic stream and body-part importance to identify influential motion cues and body regions.
\end{itemize}

\section{Method}
\label{sec:method}
As shown in Fig.~\ref{fig:overview}, our framework contains three independently trained branches designed to capture complementary motion cues and supervision signals. Branch A is a 6D rotation-based STGCN++ branch that models BVH joint rotations. Branch B is a part-aware kinetic multi-stream branch that extracts position-based cues from joint, bone, velocity, and acceleration streams with part-level attention. Branch C is a weak ambiguity-regularized kinetic branch with metadata-conditioned weak LDL. The branches are optimized separately and fused only at inference time by a probability-level ensemble.

\subsection{6D Rotation-Based STGCN++}
Branch A follows the official 6D rotation-based STGCN++ baseline~\cite{duan2022pyskl}, which models BVH joint rotations using a spatio-temporal graph convolutional network inspired by ST-GCN~\cite{yan2018stgcn}.
We first extract joint rotations from the BVH file, convert them into quaternions, and then transform them into a continuous 6D rotation representation~\cite{zhou2019rotation6d}. The resulting sequence is fed into an STGCN++ encoder, followed by a linear classifier head that produces emotion logits \(\mathbf{z}_a\).
Following the notation in Fig.~\ref{fig:overview}, the training loss for Branch A is defined as
\begin{equation}
  \mathcal{L}^{(A)} = L_a = \mathrm{CE}(\mathbf{z}_a, y),
  \label{eq:branch_a_loss}
\end{equation}
where \(\mathrm{CE}\) denotes the cross-entropy loss and \(y\) denotes the ground-truth hard emotion label. The conversion from BVH rotations to 6D representations is a fixed preprocessing step, while the STGCN++ encoder and linear classifier head are trainable.

\subsection{Part-Aware Kinetic Multi-Stream}
\noindent\textbf{Kinetic Stream Construction.}
Branch B uses 3D joint positions and explicitly constructs four kinetic streams: joint, bone, velocity, and acceleration, following multi-stream skeleton recognition methods in which joint, bone, and motion-related cues provide complementary information~\cite{shi2019twostreamagcn,shi2020msaagcn}.
The joint stream represents normalized 3D joint coordinates after root-centering and scale normalization. The bone stream represents parent-child joint displacement, capturing relative posture and local body structure~\cite{shi2019twostreamagcn}. The velocity and acceleration streams are computed as first- and second-order temporal differences of normalized joint positions, respectively.

\noindent\textbf{Stream-wise Encoding and Auxiliary Supervision.}
Each stream \(s \in \mathcal{M}\) is processed by an independent STGCN++ encoder to obtain the encoder output \(\mathbf{H}_s\), where
$\mathcal{M} =
\{\mathrm{joint}, \mathrm{bone}, \mathrm{velocity}, \mathrm{acceleration}\}$.
To encourage each stream encoder to learn discriminative emotion-related features before fusion, we attach an auxiliary classifier head directly to each encoder output \(\mathbf{H}_s\). This produces stream-specific logits \(\mathbf{z}_s\), and the multi-stream auxiliary loss \(L_m\) is defined as
\begin{equation}
  L_m = \lambda_m\sum_{s \in \mathcal{M}} \mathrm{CE}(\mathbf{z}_s, y).
  \label{eq:multi_stream_aux_loss}
\end{equation}
Here, \(\lambda_m\) is a hyperparameter that controls the contribution of the multi-stream auxiliary loss.

\noindent\textbf{Part-Attention Fusion.}
Instead of directly pooling features over the whole body, we introduce a part-attention module after each stream encoder. We group the skeleton joints into four semantic body parts: head, torso, arms, and legs, motivated by prior body-part and part-stream skeleton modeling studies~\cite{trivedi2022psumnet,long2023stepcatformer,maeda2025frequency}. For each stream, the encoder output \(\mathbf{H}_s\) is pooled within each body part to obtain part-level features. A lightweight MLP then predicts part attention weights over the four body parts, which are used to aggregate the part-level features into a part-attended stream feature \(\mathbf{f}_s\). The part-attended features from the four streams are concatenated and passed through an MLP fusion module to produce the final fused logits \(\mathbf{z}_b\). The main hard-label classification loss \(L_b\) is computed from the fused multi-stream logits:
\begin{equation}
  L_b = \mathrm{CE}(\mathbf{z}_b, y).
  \label{eq:branch_b_ce}
\end{equation}
Finally, the training loss for Branch B follows the notation in Fig.~\ref{fig:overview}:
\begin{equation}
  \mathcal{L}^{(B)} = L_b + L_m .
  \label{eq:branch_b_loss}
\end{equation}

\subsection{Intensity-Aware Weak Label Distribution Learning}
Branch C is designed to provide weak regularization for label ambiguity on top of the kinetic multi-stream backbone. Different emotion categories may share similar body motion patterns. For example, fear and surprise may involve sudden reactions, while guilt, shame, and sadness may exhibit inward or contracted body postures. To account for this ambiguity, Branch C augments hard-label classification with intensity-aware weak label distribution learning. This design builds on label distribution learning, label smoothing, and learning from soft labels~\cite{geng2016ldl,muller2019labelsmoothing,devries2024softlabels}, as well as emotion-recognition studies that treat labels as soft, uncertain, or ambiguity-aware targets~\cite{chou2020corater,mao2020emotionprofile,prabhu2024labeluncertainty,le2023uncertaintyawareldl,wu2026amber2}. Although these studies are not specific to skeleton motion, they motivate our treatment of emotion labels as weak distributions rather than only one-hot targets. Branch C uses the same joint, bone, velocity, and acceleration streams as Branch B, but adds intensity-aware weak LDL and metadata-based auxiliary supervision during training. We consider contextual metadata available in the dataset, including country, scenario ID, and scenario intensity. The metadata logits are not added to the final motion logits and are not used during inference, reducing the risk that metadata directly becomes a shortcut for classification. We do not use scenario text or actor ID, because scenario text is not assumed to be available for the hidden test set and actor ID may introduce identity-related shortcuts.

\noindent\textbf{Intensity-Based Soft Label Design.}
For Branch C, we construct an intensity-aware soft label distribution from the hard label. The soft label is defined as
\begin{equation}
  \tilde{\mathbf{y}} =
  (1-\epsilon_r)\mathbf{e}_y + \epsilon_r \mathbf{q}_y,
  \label{eq:intensity_soft_label}
\end{equation}
where \(\mathbf{e}_y\) is the one-hot vector of the ground-truth label \(y\), and \(\mathbf{q}_y\) is a coarse emotion-family distribution determined by the ground-truth class. Based on the DIEM-A label taxonomy, we divide the 12 emotion classes into two coarse groups, the basic emotion group \(\mathcal{B}\) and the social emotion group \(\mathcal{S}\). This grouping is used as a coarse heuristic for distributing ambiguity among related emotion categories:
\begin{equation}
  \begin{aligned}
    \mathcal{B}=\{&\text{anger}, \text{contempt}, \text{disgust}, \text{fear},
      \text{joy},\\
    &\text{sadness}, \text{surprise}\},\\
    \mathcal{S}=\{&\text{jealousy}, \text{shame}, \text{guilt},
    \text{gratitude}, \text{pride}\}.
  \end{aligned}
  \label{eq:emotion_groups}
\end{equation}
For each class index \(k\), \(\mathbf{q}_y\) is defined as
\begin{equation}
  q_y(k)=
  \begin{cases}
    \frac{1}{|\mathcal{B}|}, & y \in \mathcal{B},\ k \in \mathcal{B},\\
    \frac{1}{|\mathcal{S}|}, & y \in \mathcal{S},\ k \in \mathcal{S},\\
    0, & \text{otherwise}.
  \end{cases}
  \label{eq:emotion_family_distribution}
\end{equation}
Thus, \(\mathbf{q}_y\) assigns uniform probability mass to emotion classes in the same coarse emotion group as the ground-truth class \(y\), and zero probability to classes in the other group. The smoothing coefficient \(\epsilon_r\) is determined by the scenario intensity \(r\):
\begin{equation}
  \epsilon_r =
  \begin{cases}
    0.14, & r=\text{Low},\\
    0.08, & r=\text{Middle},\\
    0.04, & r=\text{High}.
  \end{cases}
  \label{eq:intensity_smoothing}
\end{equation}
The smoothing coefficients are empirically selected on validation folds, with stronger smoothing for low-intensity samples because their motion cues can be more ambiguous.

\noindent\textbf{Metadata Auxiliary Head.}
The metadata fields, namely country, scenario ID, and scenario intensity, are embedded, concatenated, and passed into a small MLP metadata auxiliary head to produce metadata logits \(\mathbf{z}_m\). These logits are not added to the final motion logits. Instead, the metadata auxiliary head is trained only through an auxiliary LDL loss:
\begin{equation}
  L_c^1 =
  \lambda_{\mathrm{meta}} \mathrm{KL}
  \left(
    \tilde{\mathbf{y}}
    \,\|\,
    \mathrm{softmax}(\mathbf{z}_m)
  \right),
  \label{eq:metadata_ldl_loss}
\end{equation}
where $\mathrm{KL}$ denotes the Kullback--Leibler divergence between a target distribution \(\mathbf{p}\) and a predicted distribution \(\mathbf{q}\). This design prevents metadata logits from directly acting as a shortcut for final classification.

\noindent\textbf{Motion-Level Supervision.}
The final motion logits of Branch C are denoted by \(\mathbf{z}_c\). We train them with both the hard label and the intensity-aware soft label:
\begin{equation}
  L_c^2 =
  \mathrm{CE}(\mathbf{z}_c, y),
  \label{eq:branch_c_ce}
\end{equation}
\begin{equation}
  L_c^3 =
  \lambda_{\mathrm{soft}}
  \mathrm{KL}
  \left(
    \tilde{\mathbf{y}}
    \,\|\,
    \mathrm{softmax}(\mathbf{z}_c)
  \right).
  \label{eq:motion_ldl_loss}
\end{equation}
In addition, Branch C uses the same multi-stream auxiliary loss \(L_m\) defined in (\ref{eq:multi_stream_aux_loss}), which is computed over the joint, bone, velocity, and acceleration streams as in Branch B.

\noindent\textbf{Overall Training Objective.}
Following the notation in Fig.~\ref{fig:overview}, the overall training loss for Branch C is
\begin{equation}
  \mathcal{L}^{(C)} = L_c^1 + L_c^2 + L_c^3 + L_m.
  \label{eq:branch_c_loss}
\end{equation}
Branch C uses scenario intensity to construct weak soft labels for modeling label ambiguity, while the metadata auxiliary head is trained separately without injecting metadata logits into the final motion logits. This reduces the risk of over-reliance on metadata shortcuts.

\subsection{Inference by Weighted Probability Ensemble}
After training, the three branches are used independently for inference. Given a test sequence, Branch A, Branch B, and Branch C produce logits \(\mathbf{z}_a\), \(\mathbf{z}_b\), and \(\mathbf{z}_c\), respectively. We first convert them into probability distributions:
\begin{equation}
  \begin{aligned}
    \mathbf{p}_a &= \mathrm{softmax}(\mathbf{z}_a), \\
    \mathbf{p}_b &= \mathrm{softmax}(\mathbf{z}_b), \\
    \mathbf{p}_c &= \mathrm{softmax}(\mathbf{z}_c).
  \end{aligned}
  \label{eq:branch_probabilities}
\end{equation}
The final prediction \(\mathbf{p}_{\mathrm{final}}\) is obtained by probability-level averaging:
\begin{equation}
  \mathbf{p}_{\mathrm{final}}
  =
  \omega_a \mathbf{p}_a +
  \omega_b \mathbf{p}_b +
  \omega_c \mathbf{p}_c,
  \label{eq:weighted_ensemble}
\end{equation}
where $\omega_a$, $\omega_b$, and $\omega_c$ denote the ensemble weights assigned to Branches A, B, and C, respectively. The predicted emotion is the class with the largest value in \(\mathbf{p}_{\mathrm{final}}\).
We emphasize that the ensemble is performed at the probability level after branch-wise softmax, rather than by directly adding logits from different branches. This avoids combining independently trained logits with potentially different scales. The probability-level ensemble allows the rotation-based branch, the part-aware kinetic branch, and the weak ambiguity-regularized kinetic branch to provide complementary predictions.

\section{Experiments}
\label{sec:experiments}

\subsection{Dataset and Evaluation Protocol}
\label{subsec:dataset_protocol}

We used the official DIEM-A challenge dataset for skeleton-based emotion recognition~\cite{mmacacii2026challenge}. Its main challenge setting and input details are summarized in Table~\ref{tab:dataset_overview}. Since the hidden-test labels are not released, all experiments reported in this paper were conducted only on the training split. For development and ablation studies, we used 10-fold leave-performer-out cross-validation, where folds are separated at the actor level. To the best of our knowledge, directly comparable published DIEM-A results under the same leave-performer-out protocol are not available. Therefore, we use the official baseline and internal variants as the primary references. We report Accuracy and Macro-F1, with Macro-F1 as the primary metric.

\subsection{Implementation Details}
\label{subsec:implementation_details}

We train the three branches described in Sec.~\ref{sec:method} independently
and fuse them only at inference time. All models are optimized using SGD with a
cosine learning-rate schedule and a weight decay of \(5\times10^{-4}\). Branch A
is trained with a learning rate of 0.20, batch size of 128, and 65 epochs.
Branches B and C are trained with a learning rate of 0.05, batch size of 16, and
80 epochs. For Branches B and C, we use mixed-precision training and gradient
accumulation with three steps. We use \(\lambda_m=0.20\),
\(\lambda_{\mathrm{soft}}=0.10\), and \(\lambda_{\mathrm{meta}}=0.30\) for the
loss terms. All models were implemented in PyTorch and trained on NVIDIA GPUs;
mixed-precision training was used for Branches B and C. The final
submission uses probability-level ensemble weights of \(\omega_a=0.30\),
\(\omega_b=0.30\), and \(\omega_c=0.40\), which are selected based on
validation performance.

\begin{table}[t]
  \centering
  \caption{DIEM-A challenge setting and evaluation protocol.}
  \footnotesize
  \begin{tabular}{@{}p{0.30\columnwidth}p{0.62\columnwidth}@{}}
    \toprule
    Category & Details \\
    \midrule
    Dataset & DIEM-A challenge dataset \\
    Task & 12-class emotion classification from BVH skeleton motion \\
    Split & 7,992/1,944 train/test sequences (only the training split is used because test labels are hidden) \\
    Skeleton & 24-joint positions for kinetic branches and 25-node rotations for Branch A \\
    Metadata & country, scenario ID, scenario intensity \\
    Protocol & 10-fold leave-performer-out cross-validation \\
    Metrics & Accuracy and Macro-F1 \\
    \bottomrule
  \end{tabular}
  \label{tab:dataset_overview}
\end{table}

\subsection{Single-Branch Results}
\label{subsec:single_branch_results}

\begin{table*}[t]
  \centering
  \caption{
    Single-branch results under 10-fold leave-performer-out cross-validation.
  }
  \label{tab:single_branch_results}
  \begin{tabular}{@{}p{0.20\textwidth}p{0.46\textwidth}cc@{}}
    \toprule
    Branch & Description & Accuracy & Macro-F1 \\
    \midrule
    Baseline
    & STGCN++ with 6D rotation
    & 0.271 $\pm$ 0.037
    & 0.252 $\pm$ 0.045 \\
    A
    & Our reproduction of the 6D rotation-based baseline
    & 0.273 $\pm$ 0.047
    & 0.252 $\pm$ 0.047 \\
    B w/o Part Attention
    & Kinetic multi-stream branch with four motion streams
    & 0.317 $\pm$ 0.051
    & 0.301 $\pm$ 0.051 \\
    B
    & Part-aware kinetic multi-stream branch with body-part encoding
    & \textbf{0.328} $\pm$ 0.045
    & 0.317 $\pm$ 0.047 \\
    C
    & Kinetic multi-stream branch with metadata-conditioned weak LDL
    & \textbf{0.328} $\pm$ 0.042
    & \textbf{0.318} $\pm$ 0.044 \\
    \bottomrule
  \end{tabular}
\end{table*}

Table~\ref{tab:single_branch_results} reports the single-branch inference results under 10-fold leave-performer-out cross-validation. Each branch is trained independently and evaluated alone, without probability-level ensemble.
Branch A achieved a Macro-F1 of 0.252, closely matching the official baseline and confirming that our reproduction provides a valid reference point for subsequent comparisons.
Introducing the kinetic multi-stream representation improved Macro-F1 from 0.252 to 0.301, suggesting that position-based motion cues such as bone vectors, velocity, and acceleration are effective for skeleton-based emotion recognition.
Adding part-level attention further improved Macro-F1 from 0.301 to 0.317, indicating that body-part-level modeling provides additional discriminative cues for emotional body motion.
Branch C achieved a Macro-F1 of 0.318, only slightly higher than Branch B's 0.317, suggesting that it provides weak ambiguity regularization rather than a substantial standalone improvement while still potentially contributing complementary predictions to the final ensemble.

\subsection{Probability-Level Ensemble Results}
\label{subsec:ensemble_results}

\begin{table}[t]
  \centering
  \caption{
    Probability-level ensemble results under 10-fold leave-performer-out
    cross-validation. Here, B denotes Branch B, the part-aware kinetic multi-stream
    branch.
  }
  \label{tab:ensemble_results}
  \footnotesize
  \setlength{\tabcolsep}{3pt}
  \begin{tabular}{@{}lccc@{}}
    \toprule
    Combination & Weights & Accuracy & Macro-F1 \\
    \midrule
    Baseline & -- & 0.271 $\pm$ 0.037 & 0.252 $\pm$ 0.045 \\
    A + B & 0.50 / 0.50 & 0.347 $\pm$ 0.056 & 0.334 $\pm$ 0.056 \\
    A + C & 0.40 / 0.60 & 0.347 $\pm$ 0.046 & 0.333 $\pm$ 0.047 \\
    B + C & 0.50 / 0.50 & 0.352 $\pm$ 0.050 & 0.341 $\pm$ 0.052 \\
    A + B + C & 0.30 / 0.30 / 0.40 & \textbf{0.366} $\pm$ 0.054 & \textbf{0.353} $\pm$ 0.055 \\
    \bottomrule
  \end{tabular}
\end{table}

\noindent\textbf{Ensemble performance.} Table~\ref{tab:ensemble_results} summarizes the probability-level ensemble results under 10-fold leave-performer-out cross-validation. Unlike Table~\ref{tab:single_branch_results}, these results combine the branch-wise softmax probabilities from multiple independently trained branches during inference.
All ensemble combinations consistently improve over the baseline. Among the two-branch ensembles, B+C achieves the highest Macro-F1, suggesting that Branch B and Branch C provide complementary predictions despite sharing a kinetic backbone. Adding Branch A to the B+C ensemble further improves performance, even though Branch A is weaker as a standalone branch. This suggests that the 6D rotation representation provides complementary information to the position-based kinetic branches.

\begin{table}[t]
  \centering
  \caption{Pairwise complementarity using out-of-fold predictions.}
  \label{tab:pairwise_complementarity}
  \footnotesize
  \begin{tabular}{@{}lcc@{}}
    \toprule
    Branch comparison & Disagreement rate & Oracle accuracy \\
    \midrule
    A vs. C & 0.734 & 0.465 \\
    A vs. B & 0.720 & 0.460 \\
    B vs. C & 0.565 & 0.452 \\
    \bottomrule
  \end{tabular}
\end{table}

\noindent\textbf{Pairwise branch complementarity.} To assess whether the branches provide complementary predictions, we examine the disagreement rate and oracle accuracy of each branch pair using out-of-fold predictions. Table~\ref{tab:pairwise_complementarity} reports this analysis. Although Branch A is weaker as a standalone model, it shows high disagreement with Branches B and C, indicating that 6D rotation cues capture information complementary to position-based kinetic cues. Branches B and C share the same kinetic backbone and therefore show lower disagreement, but their oracle accuracy still indicates complementary behavior. These results support the use of probability-level ensembling rather than selecting a single branch.

\subsection{Class- and Metadata-Wise Analysis}
\label{subsec:class_metadata_analysis}

We further analyze class-wise and metadata-related behavior using confusion pairs, intensity-wise results, and country-wise results. These analyses help identify remaining ambiguities and evaluate whether the ensemble improvement is consistent across different subsets.

\begin{table}[t]
  \centering
  \caption{Top confusion pairs of the final ensemble.}
  \label{tab:top_confusion_pairs}
  \footnotesize
  \setlength{\tabcolsep}{3pt}
  \begin{tabular}{@{}llcc@{}}
    \toprule
    True & Predicted & Count & Row rate \\
    \midrule
    jealousy & contempt & 129 & 0.194 \\
    guilt & sadness & 127 & 0.191 \\
    pride & gratitude & 122 & 0.183 \\
    disgust & fear & 108 & 0.162 \\
    fear & surprise & 95 & 0.143 \\
    surprise & fear & 95 & 0.143 \\
    joy & gratitude & 74 & 0.111 \\
    \bottomrule
  \end{tabular}
\end{table}

\noindent\textbf{Class-wise confusion.} Table~\ref{tab:top_confusion_pairs} reports the most frequent off-diagonal confusion pairs for the final ensemble. Count denotes the number of samples whose ground-truth label is the True class but were misclassified as the Predicted class. Row rate denotes the proportion of such errors among all samples of the corresponding ground-truth class. The confusion-pair analysis shows that many errors occur between semantically or behaviorally related emotions. For example, fear and surprise are mutually confused, while guilt is often predicted as sadness and pride as gratitude. These confusions suggest that emotion labels are not always separable as independent one-hot categories from body motion alone, supporting the motivation for weak label distribution learning.

\begin{table}[t]
  \centering
  \caption{Intensity-wise Macro-F1 using out-of-fold predictions.}
  \label{tab:intensity_performance}
  \footnotesize
  \setlength{\tabcolsep}{3pt}
  \begin{tabular}{@{}lccc@{}}
    \toprule
    Intensity & B & C & Ensemble \\
    \midrule
    Low & 0.286 & 0.280 & 0.308 \\
    Middle & 0.332 & 0.335 & 0.371 \\
    High & 0.348 & 0.344 & 0.383 \\
    \bottomrule
  \end{tabular}
\end{table}

\noindent\textbf{Intensity-wise performance.} The intensity-wise results in Table~\ref{tab:intensity_performance} show a clear trend: high-intensity expressions are easier to classify than middle- and low-intensity expressions. This supports the assumption that low-intensity body motion contains weaker and more ambiguous emotion cues. The final ensemble improves Macro-F1 across all intensity levels, suggesting that the ensemble gain is not limited to a single intensity condition.

\noindent\textbf{Country-wise performance.} The country-wise analysis in Table~\ref{tab:country_performance} shows that the final ensemble improves performance for both JP and TW subsets. Although TW samples show slightly higher Macro-F1, no severe performance collapse is observed for either group. This suggests that the proposed framework provides stable improvements across the two country subsets in the challenge data. We avoid making strong cultural claims because the analysis is based on challenge splits and performer distributions.

\begin{table}[H]
  \centering
  \caption{Country-wise Macro-F1 using out-of-fold predictions.}
  \label{tab:country_performance}
  \footnotesize
  \setlength{\tabcolsep}{3pt}
  \begin{tabular}{@{}lccccc@{}}
    \toprule
    Country & A & B w/o Part & B & C & Ensemble \\
    \midrule
    JP & 0.256 & 0.303 & 0.318 & 0.319 & 0.347 \\
    TW & 0.270 & 0.309 & 0.329 & 0.321 & 0.365 \\
    \bottomrule
  \end{tabular}
\end{table}

\FloatBarrier
\subsection{Explainability Ablation}
\label{subsec:explainability_ablation}

\begin{table}[H]
  \centering
  \caption{
    Explainability ablation results using Branch B.
  }
  \footnotesize
  \begin{tabular}{@{}lcc@{}}
    \toprule
    Condition & Macro-F1 & Drop \\
    \midrule
    \multicolumn{3}{@{}l}{\textit{(a) Kinematic stream ablation}} \\
    Full branch & \textbf{0.323} & \textbf{0.000} \\
    w/o velocity & 0.209 & 0.114 \\
    w/o bone & 0.211 & 0.112 \\
    w/o acceleration & 0.273 & 0.051 \\
    w/o joint & 0.278 & 0.045 \\
    \midrule
    \multicolumn{3}{@{}l}{\textit{(b) Body-part ablation}} \\
    Full branch & \textbf{0.323} & \textbf{0.000} \\
    w/o arms & 0.058 & 0.265 \\
    w/o legs & 0.156 & 0.167 \\
    w/o head & 0.281 & 0.042 \\
    w/o torso & 0.306 & 0.017 \\
    \bottomrule
  \end{tabular}
  \label{tab:explainability_ablation}
\end{table}

Table~\ref{tab:explainability_ablation} summarizes explainability ablations using Branch B. Macro-F1 is computed on pooled out-of-fold predictions over all training samples, while Table~\ref{tab:single_branch_results} reports the mean and standard deviation of fold-wise Macro-F1. A larger Macro-F1 drop indicates a larger contribution of the removed component. Among the kinematic stream ablations, removing either the velocity stream or the bone stream individually causes the largest drops, indicating that temporal dynamics and parent-child joint relationships are important for emotion recognition. Among the body-part ablations, removing either the arms or the legs individually leads to the largest drops, suggesting that dynamic limb movements are especially informative for distinguishing emotional body expressions.

\FloatBarrier
\section{Conclusion}
\label{sec:conclusion}

This paper presented a multi-branch skeleton-based emotion recognition framework for the DIEM-A task in the MMAC ACII 2026 Challenge. The proposed approach combines a 6D rotation-based branch, a part-aware kinetic multi-stream branch, and a metadata-conditioned weak label distribution learning branch through probability-level ensemble. Experimental results under 10-fold leave-performer-out cross-validation demonstrate consistent improvements over the official baseline, achieving the best performance with an Accuracy of 0.366 and a Macro-F1 of 0.353. Explainability analyses further suggest that velocity and bone streams, together with arm and leg movements, provide important cues for recognizing emotional body motion. The proposed system is designed as a practical challenge solution and has several limitations. Evaluation is limited to the official training data because hidden test labels are unavailable, and the contributions of individual components in Branch C are not fully isolated. Moreover, the multi-branch ensemble increases inference cost. Future work will investigate finer-grained component ablations, more efficient single-model architectures, and broader evaluation on diverse datasets.

\section*{ACKNOWLEDGMENT}
This work was supported by JSPS KAKENHI Grant Number JP26K02785 and JST CREST Grant Number JPMJCR2554.

\section*{Ethical Impact Statement}
This work uses the official DIEM-A challenge dataset and does not collect new human-subject data. Although BVH skeleton motion avoids directly using facial appearance or speech content, emotion recognition from body motion can still raise privacy and misuse risks if deployed without consent. Such systems should not be used for surveillance, employment evaluation, education assessment, or other consequential decisions without explicit consent, human oversight, and careful validation.

The findings are limited to the DIEM-A challenge setting, which contains Japanese and Taiwanese performers and predefined basic and social emotion categories. Therefore, the model should not be assumed to generalize to other cultures, populations, contexts, or naturalistic emotional behavior. Metadata is used only during training and metadata logits are not used at inference time, reducing the risk of direct metadata shortcuts. Future work should further examine bias, fairness, and robustness across groups.

\bibliographystyle{IEEEtran}
\bibliography{references}

@IEEEtranBSTCTL{IEEEtranBSTCTL:disable_dash,
  CTLdash_repeated_names = "no"
}

@misc{mmacacii2026challenge,
  author       = {{MMAC@ACII 2026 Organizers}},
  title        = {{MMAC@ACII 2026 Challenge}},
  howpublished = {\url{https://sites.google.com/view/mmac-acii-2026/overview?authuser=0}},
  year         = {2026},
  note         = {Accessed: Jun. 10, 2026}
}

@inproceedings{cheng2025diema,
  author    = {Cheng, Miao and Tseng, Chia-huei and Fujiwara, Ken and Schneider, Victor and Kitamura, Yoshifumi},
  title = {Asian Emotional Body Movement Database: {Diverse Intercultural E-Motion Database of Asian Performers} ({DIEM-A})},
  booktitle = {Proceedings of the 13th International Conference on Affective Computing and Intelligent Interaction ({ACII})},
  year      = {2025}
}

@article{duan2022pyskl,
  author  = {Duan, Haodong and Wang, Jiaqi and Chen, Kai and Lin, Dahua},
  title   = {{PYSKL}: Towards Good Practices for Skeleton Action Recognition},
  journal = {arXiv preprint arXiv:2205.09443},
  year    = {2022}
}

@inproceedings{yan2018stgcn,
  author    = {Yan, Sijie and Xiong, Yuanjun and Lin, Dahua},
  title     = {Spatial Temporal Graph Convolutional Networks for Skeleton-Based Action Recognition},
  booktitle = {Proceedings of the AAAI Conference on Artificial Intelligence},
  year      = {2018}
}

@inproceedings{shi2019twostreamagcn,
  author    = {Shi, Lei and Zhang, Yifan and Cheng, Jian and Lu, Hanqing},
  title     = {Two-Stream Adaptive Graph Convolutional Networks for Skeleton-Based Action Recognition},
  booktitle = {Proceedings of the IEEE/CVF Conference on Computer Vision and Pattern Recognition ({CVPR})},
  year      = {2019}
}

@article{shi2020msaagcn,
  title={Skeleton-based action recognition with multi-stream adaptive graph convolutional networks},
  author={Shi, Lei and Zhang, Yifan and Cheng, Jian and Lu, Hanqing},
  journal={IEEE Transactions on Image Processing},
  volume={29},
  pages={9532--9545},
  year={2020},
  publisher={IEEE}
}

@inproceedings{zhou2019rotation6d,
  author    = {Zhou, Yi and Barnes, Connelly and Lu, Jingwan and Yang, Jimei and Li, Hao},
  title     = {On the Continuity of Rotation Representations in Neural Networks},
  booktitle = {Proceedings of the IEEE/CVF Conference on Computer Vision and Pattern Recognition ({CVPR})},
  year      = {2019}
}

@article{geng2016ldl,
  author  = {Geng, Xin},
  title   = {Label Distribution Learning},
  journal = {IEEE Transactions on Knowledge and Data Engineering},
  volume  = {28},
  number  = {7},
  pages   = {1734--1748},
  year    = {2016}
}

@inproceedings{le2023uncertaintyawareldl,
  author    = {Le, Nhat and others},
  title     = {Uncertainty-Aware Label Distribution Learning for Facial Expression Recognition},
  booktitle = {Proceedings of the IEEE/CVF Winter Conference on Applications of Computer Vision ({WACV})},
  pages     = {6088--6097},
  year      = {2023}
}

@article{wu2026amber2,
  author  = {Wu, Jingyao and Lin, Grace and Song, Yinuo and Picard, Rosalind},
  title   = {{AmbER$^2$}: Dual Ambiguity-Aware Emotion Recognition Applied to Speech and Text},
  journal = {arXiv preprint arXiv:2601.18010},
  year    = {2026}
}

@inproceedings{chou2020corater,
  author    = {Chou, Huang-Cheng and Lee, Chi-Chun},
  title     = {Learning to Recognize Per-Rater's Emotion Perception Using Co-Rater Training Strategy with Soft and Hard Labels},
  booktitle = {Proceedings of Interspeech 2020},
  pages     = {4108--4112},
  year      = {2020},
  doi       = {10.21437/Interspeech.2020-1714}
}

@inproceedings{mao2020emotionprofile,
  author    = {Mao, Shuiyang and Ching, P. C. and Lee, Tan},
  title     = {Emotion Profile Refinery for Speech Emotion Classification},
  booktitle = {Proceedings of Interspeech 2020},
  pages     = {531--535},
  year      = {2020},
  doi       = {10.21437/Interspeech.2020-1771}
}

@article{prabhu2024labeluncertainty,
  author  = {Prabhu, Navin Raj and Lehmann-Willenbrock, Nale and Gerkmann, Timo},
  title   = {End-to-End Label Uncertainty Modeling in Speech Emotion Recognition Using Bayesian Neural Networks and Label Distribution Learning},
  journal = {IEEE Transactions on Affective Computing},
  volume  = {15},
  number  = {2},
  pages   = {579--592},
  year    = {2024},
  doi     = {10.1109/TAFFC.2023.3283595}
}

@inproceedings{muller2019labelsmoothing,
  author    = {M{\"u}ller, Rafael and Kornblith, Simon and Hinton, Geoffrey E.},
  title     = {When Does Label Smoothing Help?},
  booktitle = {Advances in Neural Information Processing Systems},
  volume    = {32},
  year      = {2019}
}

@article{noroozi2018emotionalbodygesture,
  author  = {Noroozi, Fatemeh and Corneanu, Ciprian Adrian and Kami{\'n}ska, Dorota and Sapi{\'n}ski, Tomasz and Escalera, Sergio and Anbarjafari, Gholamreza},
  title   = {Survey on Emotional Body Gesture Recognition},
  journal = {arXiv preprint arXiv:1801.07481},
  year    = {2018}
}

@article{lu2025skeletonemotion,
  author  = {Lu, Haifeng and Chen, Jiuyi and Zhang, Zhen and Liu, Ruida and Zeng, Runhao and Hu, Xiping},
  title   = {Emotion Recognition from Skeleton Data: A Comprehensive Survey},
  journal = {arXiv preprint arXiv:2507.18026},
  year    = {2025}
}

@article{piana2014bodygestures,
  author  = {Piana, Stefano and Staglian{\`o}, Alessandra and Odone, Francesca and Verri, Alessandro and Camurri, Antonio},
  title   = {Real-time Automatic Emotion Recognition from Body Gestures},
  journal = {arXiv preprint arXiv:1402.5047},
  year    = {2014}
}

@article{devries2024softlabels,
  author  = {de Vries, Sjoerd and Thierens, Dirk},
  title   = {Learning with Confidence: Training Better Classifiers from Soft Labels},
  journal = {arXiv preprint arXiv:2409.16071},
  year    = {2024}
}

@article{trivedi2022psumnet,
  author  = {Trivedi, Neel and Sarvadevabhatla, Ravi Kiran},
  title   = {{PSUMNet}: Unified Modality Part Streams are All You Need for Efficient Pose-Based Action Recognition},
  journal = {arXiv preprint arXiv:2208.05775},
  year    = {2022}
}

@article{long2023stepcatformer,
  author  = {Long, Nguyen Huu Bao},
  title   = {{STEP CATFormer}: Spatial-Temporal Effective Body-Part Cross Attention Transformer for Skeleton-Based Action Recognition},
  journal = {arXiv preprint arXiv:2312.03288},
  year    = {2023}
}

@article{maeda2025frequency,
  title={Frequency-guided multi-level human action anomaly detection with normalizing flows},
  author={Maeda, Shun and Gu, Chunzhi and Yu, Jun and Tokai, Shogo and Gao, Shangce and Zhang, Chao},
  journal={Pattern Recognition},
  pages={112770},
  year={2025},
  publisher={Elsevier}
}

@inproceedings{wei2026human,
  title     = {Human Action Recognition via Dataset Condensation},
  author    = {Wei, Yijin and Zou, Yawen and Kamide, Koichiro and Gu, Chunzhi and Ge, Hangli and Zhang, Chao},
  booktitle = {Proceedings of NICOGRAPH International 2026},
  year      = {2026}
}

@article{kamide2025few,
  title={Few-shot human action anomaly detection via a unified contrastive learning framework},
  author={Kamide, Koichiro and Sakai, Shunsuke and Maeda, Shun and Gu, Chunzhi and Zhang, Chao},
  journal={Knowledge-Based Systems},
  pages={115133},
  year={2025},
  publisher={Elsevier}
}

@article{gu2024orientation,
  title={Orientation-aware leg movement learning for action-driven human motion prediction},
  author={Gu, Chunzhi and Zhang, Chao and Kuriyama, Shigeru},
  journal={Pattern Recognition},
  volume={150},
  pages={110317},
  year={2024},
  publisher={Elsevier}
}

\end{document}